\documentclass[twoside]{article} 

\pdfoutput=1

\usepackage[final, nonatbib]{nips_2016}
\usepackage{times}
\usepackage{helvet}
\usepackage{courier}

\usepackage[utf8]{inputenc} 
\usepackage[T1]{fontenc} 
\usepackage{url} 
\usepackage{booktabs} 
\usepackage{amsfonts} 
\usepackage{nicefrac} 
\usepackage{microtype} 
\usepackage{amsmath}
\usepackage{algorithmicx}
\usepackage{algpseudocode}

\usepackage{color}

\usepackage{graphicx}
\usepackage{epstopdf} 
\usepackage{ifthen}

\newcommand{\beq}[1][]{\begin{equation}}
\newcommand{\eeq}{\end{equation}}
\newcommand{\beqr}{\begin{eqnarray}}
\newcommand{\eeqr}{\end{eqnarray}}

\newcommand{\req}[1]{(\ref{#1})}

\newcommand{\kl}[2]{\textrm{KL}\left(#1||#2\right)}

\title{Hierarchy through Composition with Linearly Solvable Markov Decision Processes}
\author{
  Andrew M. Saxe*, Adam Earle$\dagger$, Benjamin Rosman$\dagger \ddagger$ \\
  *Center for Brain Science, Harvard University\\
  $\dagger$ School of Computer Science and Applied Mathematics, University of the Witwatersrand\\
  $\ddagger$Council for Scientific and Industrial Research, Pretoria, South Africa\\
  \texttt{asaxe@fas.harvard.edu, adam.earle@students.wits.ac.za},\\ 
  \texttt{ brosman@csir.co.za}
  }

\begin{document}

\maketitle

\begin{abstract}

Hierarchical architectures are critical to the scalability of reinforcement learning methods. Current hierarchical frameworks execute actions serially, with macro-actions comprising sequences of primitive actions. We propose a novel alternative to these control hierarchies based on concurrent execution of many actions in parallel. Our scheme uses the concurrent compositionality provided by the linearly solvable Markov decision process (LMDP) framework, which naturally enables a learning agent to draw on several macro-actions simultaneously to solve new tasks. We introduce the Multitask LMDP module, which maintains a parallel distributed representation of tasks and may be stacked to form deep hierarchies abstracted in space and time.

\end{abstract}

\section{Introduction}

Real world tasks unfold at a range of spatial and temporal scales, such that learning solely at the finest scale is likely to be slow. Hierarchical reinforcement learning (HRL) \cite{Barto2003,Parr1998,Dietterich2000} attempts to remedy this by learning a nested sequence of ever more detailed plans. Hierarchical schemes have a number of desirable properties. Firstly, they are intuitive, as humans seldom plan at the level of raw actions, typically preferring to reason at a higher level of abstraction \cite{Botvinick2009,Ribas-Fernandes2011,Solway2014}. Secondly, they provide computational benefits in the form of knowledge that can be reused across tasks, domains, or even between different agents, resulting in a powerful tool for multitask learning. 

Most HRL schemes rely on a serial call/return procedure, in which temporally extended macro-actions or `options' can call other (macro-)actions. In this sense, these schemes draw on a computer metaphor, in which a serial processor chooses sequential primitive actions, occasionally pushing or popping new macro-actions onto a stack. The influential options framework \cite{Sutton1999}, MAXQ method \cite{Dietterich2000,Jonssona} and the hierarchy of abstract machines \cite{Parr1998,Burridge1999} all share this sequential-execution structure. 

In this paper, we develop a novel hierarchical scheme with a fundamentally parallel and distributed execution structure by exploiting the compositionality afforded by the linearly solvable Markov decision process (LMDP) framework \cite{Todorov2006Linearly-solvableProblems,Kappen2005}. A higher-level action in our scheme is a weighted blend of all lower level tasks, all of which operate concurrently. 

We present a Multitask LMDP module that leverages this concurrent blending of behaviors for powerful multitask learning. In standard schemes, if a robot arm has acquired policies to individually reach two points in some space, this knowledge typically does not aid it in optimally reaching to either point. However in our scheme, such behaviours can be expressed as different concurrent task blends, such that an agent can simultaneously draw on several sub-policies to optimally achieve goals not explicitly represented by any of the individual behaviours, including tasks the agent has never performed before.

Next we show how this Multitask LMDP module can be stacked, resulting in a hierarchy of more abstract modules that communicate distributed representations of tasks between layers. We give a simple theoretical analysis showing that hierarchy can yield qualitative efficiency improvements, and demonstrate the operation of the method on a navigation domain.

Our scheme builds on a variety of prior work. Like the options framework \cite{Sutton1999}, we build a hierarchy in time. Similar to MAXQ Value Decomposition \cite{Dietterich2000}, we decompose a target MDP into a hierarchy of smaller SMDPs which progressively abstract over states.  From Feudal RL, we draw the idea of a managerial hierarchy in which higher layers prescribe goals but not details for lower layers \cite{Dayan1993FeudalLearning}. Most closely related to our scheme, \cite{Jonssona} develop a MAXQ decomposition within the LMDP formalism. Our method differs from all of the above approaches in permitting a concurrent blend of tasks at each level, and developing a uniform, stackable module capable of performing a variety of tasks.

\section{The Multitask LMDP: A compositional action module}

Our goal in this paper is to describe a flexible action-selection module which can be stacked to form a hierarchy, such that the full action at any given point in time is composed of the concurrent composition of sub-actions within sub-actions. By analogy to perceptual deep networks, restricted Boltzmann machines (RBMs) form a component module from which a deep belief network can be constructed by layerwise stacking \cite{Hinton2006,Hinton2006a}. We seek a similar module in the context of action or control. This section describes the module, the Multitask LMDP (MLMDP), before turning to how it can be stacked. Our formulation relies on the linearly solvable Markov decision process (LMDP) framework introduced by Todorov \cite{Todorov2006Linearly-solvableProblems} (see also \cite{Kappen2005}). The LMDP differs from the standard MDP formulation in fundamental ways, and enjoys a number of special properties. We first briefly describe the canonical MDP formulation, in order to explain what the switch to the LMDP accomplishes and why it is necessary.

\subsection{Canonical MDPs}
In its standard formulation, an MDP is a four-tuple $M=\langle S,A,P,R\rangle$, where $S$ is a set of states, $A$ is a set of discrete actions, $P$ is a transition probability distribution $P: S \times A \times S \rightarrow [0,1]$, and $R$ is an expected instantaneous reward function $R: S \times A \rightarrow \mathbb R$. The goal is to determine an optimal policy $\pi: S \rightarrow A$ specifying which action to take in each state. This optimal policy can be computed from the optimal value function $V: S \rightarrow \mathbb R$, defined as the expected reward starting in a given state and acting optimally thereafter. The value function obeys the well-known Bellman optimality condition
\beq
	V(s) = \max_{a\in A} \left\{ R(s,a) + \sum_{s^\prime} P(s^\prime | s,a)V(s^\prime) \right\}. \label{typical_bellman}
\eeq
This formalism is the basis of most practical and theoretical studies of decision-making under uncertainty and reinforcement learning \cite{Sutton1998}. See, for instance, \cite{Mnih2015,Lillicrap2015,Levine2016End-to-EndPolicies} for recent successes in challenging domains.

For the purposes of a compositional hierarchy of actions, this formulation presents two key difficulties.
\begin{enumerate}
	\item \textbf{Mutually exclusive sequential actions} First, the agent's actions are discrete and execute serially. Exactly one (macro-)action operates at any given time point. Hence there is no way to build up an action at a single time point out of several `subactions' taken in parallel. For example, a control signal for a robotic arm cannot be composed of a control decision for the elbow joint, a control decision for the shoulder joint, and a control decision for the gripper, each taken in parallel and combined into a complete action for a specific time point. 
	
	\item \textbf{Non-composable optimal policies} The maximization in Eqn.~\req{typical_bellman} over a discrete set of actions is nonlinear. This means that optimal solutions, in general, do not compose in a simple way. Consider two standard MDPs $M_1=\langle S, A, P, R_1\rangle$ and $M_2=\langle S, A, P, R_2\rangle$ which have identical state spaces, action sets, and transition dynamics but differ in their instantaneous rewards $R_1$ and $R_2$. These may be solved independently to yield value functions $V_1$ and $V_2$. Unfortunately, the value function of the MDP $M_{1+2}=\langle S, A, P, R_1+R_2 \rangle$, whose instantaneous rewards are the sum of the first two, is not $V_{1+2}=V_1 + V_2$. In general, there is no simple procedure for deriving $V_{1+2}$ from $V_1$ and $V_2$; it may only be found by solving Eqn. \req{typical_bellman} again.
\end{enumerate}

\subsection{Linearly Solvable MDPs}
By contrast, the LMDP \cite{Todorov2009,Dvijotham2011AControl,Todorov2009b,Dvijotham2010} is defined by a three-tuple $L=\langle S, P, R\rangle$, where $S$ is a set of states, $P$ is a passive transition probability distribution $P: S \times S \rightarrow [0,1]$, and $R$ is an expected instantaneous reward function $R: S \rightarrow \mathbb R$. The LMDP framework replaces the traditional discrete set of actions $A$ with a continuous probability distribution over next states $a: S \times S \rightarrow [0,1]$. That is, the `control' or `action' chosen by the agent in state $s$ is a transition probability distribution over next states, $a(\cdot|s)$. The controlled transition distribution may be interpreted either as directly constituting the agent's dynamics, or as a stochastic policy over deterministic actions which effect state transitions \cite{Todorov2006Linearly-solvableProblems,Jonssona}. Swapping a discrete action space for a continuous action space is a key change which will allow for concurrently selected `subactions' and distributed representations. 

The LMDP framework additionally requires a specific form for the cost function to be optimized. The instantaneous reward for taking action $a(\cdot|s)$ in state $s$ is
\beq
	\mathcal R(s,a) = R(s) - \lambda \kl{a(\cdot|s)}{P(\cdot|s)}, \label{inst_rew}
\eeq
where the KL term is the Kullback-Leibler divergence between the selected control transition probability and the passive dynamics. 
This term implements a control cost, encouraging actions to conform to the natural passive dynamics of a domain. 
In a cart-pole balancing task, for instance, the passive dynamics might encode the transition structure arising from physics in the absence of control input. Any deviation from these dynamics will require energy input. In more abstract settings, such as navigation in a 2D grid world, the passive dynamics might encode a random walk, expressing the fact that actions cannot transition directly to a far away goal but only move some limited distance in a specific direction. The parameter $\lambda$ in Eqn. (\ref{inst_rew}) acts to trade-off the relative value between the reward of being in a state and the control cost, and determines the stochasticity of the resulting policies.


We consider first-exit problems (see \cite{Dvijotham2011AControl} for infinite horizon and other formulations), in which the state space is divided into a set of absorbing \textit{boundary} states $\mathcal B \subset S$ and non-absorbing \textit{interior} states $\mathcal I \subset S$, with $S = \mathcal B \cup \mathcal I$. In this formulation, an agent acts in a variable length episode that consists of a series of transitions through interior states before a final transition to a boundary state which terminates the episode. The goal is to find the policy $a^*$ which maximizes the total expected reward
across the episode,
\beq
	a^*= \textrm{argmax}_{a} \mathbb E_{\stackrel{s_{t+1}\sim a(\cdot|s_t)}{\tau=\min\left\{t:s_t \in \mathcal B\right\}}} \left\{ \sum_{t=1}^{\tau-1} \mathcal R(s_t,a) + R(s_\tau) \right\}.
\eeq

Because of the carefully chosen structure of the reward $\mathcal R(s,a)$ and the continuous action space, the Bellman equation simplifies greatly. In particular define the \textit{desirability} function $z(s)=e^{V(s)/\lambda}$ as the exponentiated cost-to-go function, and define $q(s)=e^{R(s)/\lambda}$ to be the exponentiated instantaneous rewards. Let $N$ be the number of states, and $N_i$ and $N_b$ be the number of internal and boundary states respectively. Represent $z(s)$ and $q(s)$ with $N$-dimensional column vectors $z$ and $q$, and the transition dynamics $P(s^\prime|s)$ with the $N$-by-$N_i$ matrix $P$, where column index corresponds to $s$ and row index corresponds to $s^\prime$. Let $z_i$ and $z_b$ denote the partition of $z$ into internal and boundary states, respectively, and similarly for $q_i$ and $q_b$. Finally, let $P_i$ denote the $N_i$-by-$N_i$ submatrix of $P$ containing transitions between internal states, and $P_b$ denote the $N_b$-by-$N_i$ submatrix of $P$ containing transitions from internal states to boundary states.

As shown in \cite{Todorov2009b}, the Bellman equation in this setting reduces to
\beq
	(I - M_iP^T_i)z_i = M_iP^T_bz_b \label{lmdp_bellman}
\eeq
where $M_i=\textrm{diag}(q_i)$ and, because boundary states are absorbing, $z_b=q_b$. The exponentiated Bellman equation is hence a linear system, the key advantage of the LMDP framework. A variety of special properties flow from the linearity of the Bellman equation, which we exploit in the following.

Solving for $z_i$ may be done explicitly as $z_i = (I - M_iP^T_i)^{-1}M_iP^T_bz_b $ or via the z-iteration method (akin to value iteration),
\beq
	z_i \leftarrow M_iP^T_iz_i + M_iP^T_bz_b.
	\label{eqn_power_iteration}
\eeq

Finally, the optimal policy may be computed in closed form as
\beq
	a^*(s^\prime|s) = \frac{P(s^\prime|s)z(s^\prime)}{\mathcal G[z](s)},
	\label{eqn_optimal_policy}
\eeq
where the normalizing constant $\mathcal G[z](s)= \sum_{s^\prime} P(s^\prime|s)z(s^\prime)$. Detailed derivations of these results are given in \cite{Todorov2009b,Todorov2009,Dvijotham2011AControl}. Intuitively, the hard maximization of Eqn.~\req{typical_bellman} has been replaced by a soft maximization $\log(\sum \exp(\cdot))$, and the continuous action space enables closed form computation of the optimal policy.

\begin{figure*}
\begin{center}
\includegraphics[width=\textwidth]{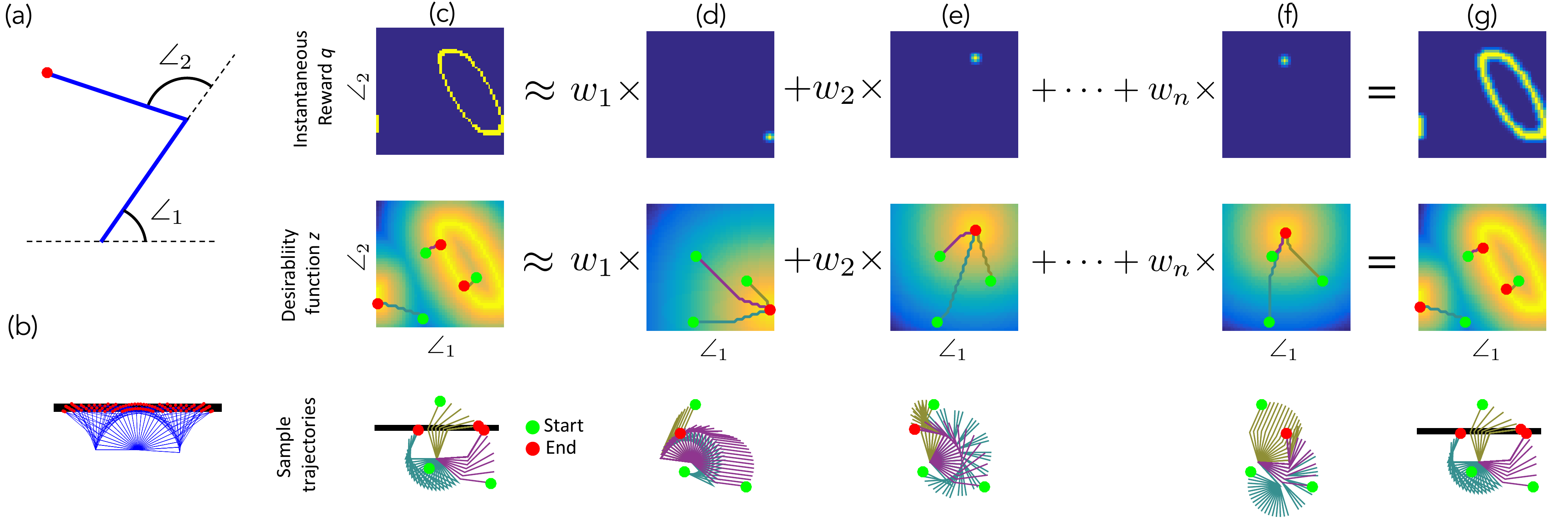}
\end{center}

\caption{Distributed task representations with the Multitask LMDP. (a) Example 2DOF arm constrained to the plane with state space consisting of shoulder and elbow joint angles $\angle_1,\angle_2\in [-\pi,\pi]$ respectively. (b) A novel task is specified as an instantaneous reward function over terminal states. In this example, the task ``reach to the black rectangle'' is encoded by rewarding any terminal state with the end effector in black rectangle (all successful configurations shown). (c-g) Solutions via the LMDP. Top row: Instantaneous rewards in the space of joint angles. Middle row: Optimal desirability function with sample trajectories starting at green circles, finishing at red circles. Bottom row: Strobe plot of sample trajectories in Cartesian space. Trajectories start at green circle, end at red circle. Column (c): The linear Bellman equation is solved for this particular instantaneous boundary reward structure to obtain the optimal value function. Columns (d-g): Solution via compositional Multitask LMDP. The instantaneous reward structure is expressed as a weighted combination of previously-learned subtasks (e-f), here chosen to be navigation to specific points in state space. (g) Because of the linearity of the Bellman equation, the resulting combined value function is optimal, and the system can act instantly despite no explicit training on the reach-to-rectangle task. The same fixed basis set can be used to express a wide variety of tasks (reach to a cross; reach to a circle; etc).}
\label{fig_distributed_task_rep}
\end{figure*}

Compared to the standard MDP formulation, the LMDP has
\begin{enumerate}
	\item \textbf{Continuous concurrent actions}  Actions are expressed as transition probabilities over next states, such that these transition probabilities can be influenced by many subtasks operating in parallel, and in a graded fashion.
		\item \textbf{Compositional optimal policies} In the LMDP, linearly blending desirability functions yields the correct composite desirability function \cite{Todorov2009b,Todorov2009}. In particular, consider two LMDPs $L_1=\langle S, P, q_i, q_b^1\rangle$ and $L_2=\langle S, P, q_i, q_b^2\rangle$ which have identical state spaces, transition dynamics, and internal reward structures, but differ in their exponentiated boundary rewards $q_b^1$ and $q_b^2$. These may be solved independently to yield desirability functions $z^1$ and $z^2$. The desirability function of the LMDP $L_{1+2}=\langle S, P, q_i, \alpha q_b^1+\beta q_b^2 \rangle$, whose instantaneous rewards are a weighted sum of the first two, is simply $z^{1+2}=\alpha z^1 + \beta z^2$. This remarkable property follows from the linearity of Eqn.~\ref{lmdp_bellman}, and is the foundation of our hierarchical scheme.
	\end{enumerate}

\subsection{The Multitask LMDP}

To build a multitask action module, we exploit this compositionality as follows: suppose that we learn a set of LMDPs $L_t=\langle S, P, q_i, q_b^t\rangle$, $t=1,\cdots,N_t$ which all share the same state space, passive dynamics, and internal rewards, but differ in their instantaneous exponentiated boundary reward structure $q^t_b, ~ t = 1, \cdots, N_t$. Each of these LMDPs corresponds to a different task, defined by its boundary reward structure, in the same overall state space (see \cite{Barreto2016} for a related approach in traditional MDPs). We denote the Multitask LMDP module $M$ formed from these $N_t$ LMDPs as $M=\langle S, P, q_i, Q_b\rangle$. Here the $N_b$-by-$N_t$ task basis matrix $Q_b=\left[ q^1_b ~ q^2_b ~ \cdots ~ q^{N_t}_b \right]$ encodes all component tasks in the MLMDP. Solving each component LMDP yields the corresponding desirability functions $z^t_i,~t=1,\cdots,N_t$ for each task, which can also be formed into the $N_i$-by-$N_t$ desirability basis matrix $Z_i = \left[z^1_i~z^2_i~\cdots~z^{N_t}_i\right]$ for the multitask module. 

When we encounter a new task defined by a novel instantaneous exponentiated reward structure $q$, if we can express it as a linear combination of previously learned tasks, $q = Q_bw$, 
where $w\in R^{N_t}$ is a vector of task blend weights, then we can instantaneously derive its optimal desirability function as $z_i = Z_iw$. This immediately yields the optimal action through Eqn.~\ref{eqn_optimal_policy}. Hence an MLMDP agent can act optimally even in never-before-seen tasks, provided that the new task lies in the subspace spanned by previously learned tasks.

More generally, if the target task $q$ is not an exact linear combination of previously learned tasks, an approximate task weighting $w$ can be found as
\begin{equation}
\textrm{argmin}_w \left\|q -  Qw\right\|\quad\textrm{subject to}\quad Qw\geq0. \label{multitask_blend}
\end{equation}
The technical requirement $Qw\geq0$ is due to the relationship $q = \textrm{exp}(r/\lambda)$, such that negative values of $q$ are not possible. In practice this can be approximately solved as $w=Q^{\dagger} q$, where $\dagger$ denotes the pseudoinverse, and negative elements of $w$ are projected back to zero.

 Just as the activation values of a bank of Gabor filters constitute a distributed representation of an image, the coefficients of the task blend $w$ constitute a distributed representation of the current task to be performed.  Although the set of basis tasks $Q_b$ is fixed and finite, they nevertheless permit an infinite space of tasks to be performed through their concurrent linear composition. Figure \ref{fig_distributed_task_rep} demonstrates this ability of the Multitask LMDP module in the context of a 2D robot arm reaching task. From knowledge of how to reach individual points in space, the module can instantly act optimally to reach to a rectangular region.

\section{Stacking the module: Concurrent Hierarchical LMDPs}

To build a hierarchy out of this Multitask LMDP module, we construct a stack of MLMDPs in which higher levels select the instantaneous reward structure that defines the current task for lower levels. To take a navigation example, a high level module might specify that the lower level module should reach room A but not B by placing instantaneous rewards in room A but no rewards in room B. Crucially, the fine details of achieving this subgoal can be left to the low-level module.
 Critical to the success of this hierarchical scheme is the flexible, optimal composition afforded by the Multitask LMDP module: the specific reward structure commanded by the higher layer will often be novel for the lower level, but will still be performed optimally provided it lies in the basis of learned tasks. We now describe this scheme formally. First, we describe how a hierarchy of MLMDPs can be derived. Next, we turn to the bidirectional flow of information between layers of the hierarchy.

\begin{figure*}
\includegraphics[width = \textwidth]{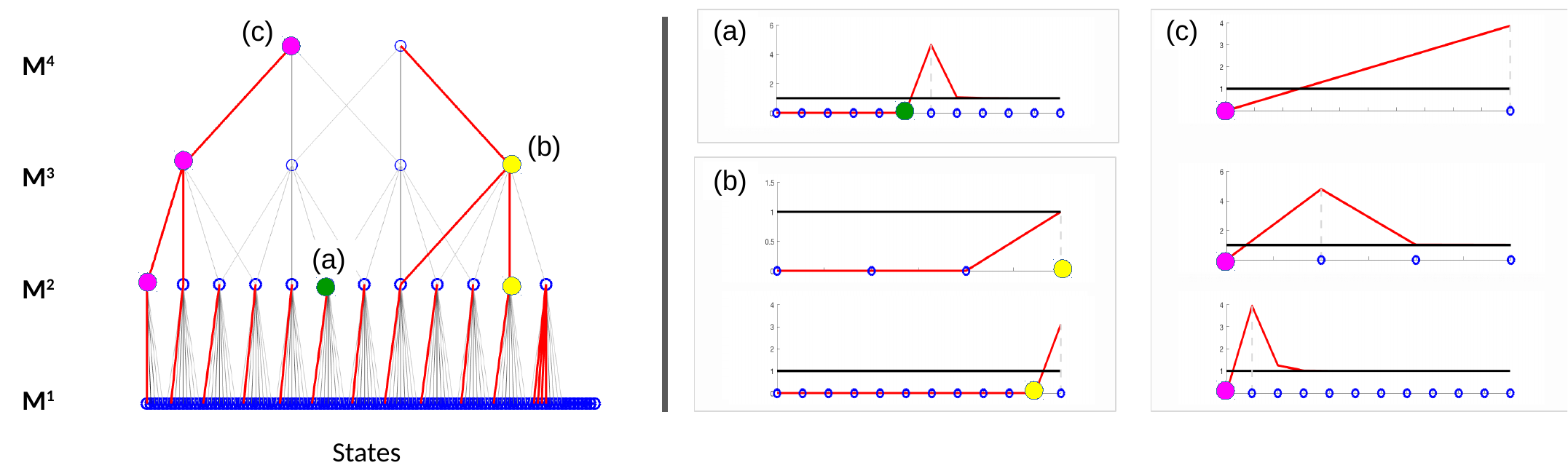}

\caption{ Stacking Multitask LMDPs. Left: Deep hierarchy for navigation through a 1D corridor. Lower-level MLMDPs are abstracted to form higher-level MLMDPs by choosing a set of `subtask' states which can be accessed by the lower level (grey lines between levels depict passive subtask transitions $P^l_t$). Lower levels access these subtask states to indicate completion of a subgoal and to request more information from higher levels; higher levels communicate new subtask state instantaneous rewards, and hence the concurrent task blend, to the lower levels. Red lines indicate higher level access points for one sample trajectory starting from leftmost state and terminating at rightmost state. Right: Panels (a-c) depict distributed task blends arising from accessing the hierarchy at points denoted in left panel. The higher layer states accessed are indicated by filled circles. (a) Just the second layer of hierarchy is accessed, resulting in higher weight on the task to achieve the next subgoal and zero weights on already achieved subgoals. (b) The second and third levels are accessed, yielding new task blends for both. (c) All levels are accessed yielding task blends at a range of scales.}
\label{hierarchy_overview}

\end{figure*}

\subsection{Constructing a hierarchy of MLMDPs}

We start with the MLMDP $M^1=\left\langle S^1, P^{1}, q^{1}_i, Q^{1}_b\right\rangle$ that we must solve, where here the superscript denotes the hierarchy level (Fig.~\ref{hierarchy_overview}). This serves as the base case for our recursive scheme for generating a hierarchy. For the inductive step, given an MLMDP $M^{l}=\left\langle S^{l},P^{l}, q^{l}_i, Q^{l}_b\right\rangle$ at level $l$, we augment the state space $\tilde S^l=S^l \cup S_t^l$ with a set of $N_t$ terminal boundary states $S_t^l$ that we call \textit{subtask} states. Semantically, entering one of these states will correspond to a decision by the layer $l$ MLMDP to access the next level of the hierarchy. The transitions to subtask states are governed by a new $N^{l}_t$-by-$N^{l}_i$ passive dynamics matrix $P^{l}_t$, which is chosen by the designer to encode the structure of the domain. By choosing $N_t^l<N_i^l$, this will yield state abstraction. In the augmented MLMDP, the passive dynamics become $\tilde P^l = \mathcal{N}(\left[P_i^l;~ P_b^{l}; ~ P_t^{l}\right])$ where the operation $\mathcal{N}(\cdot)$ renormalizes each column to sum to one, making it a valid transition distribution.

It remains to specify the matrix of subtask-state instantaneous rewards $R_t^l$ for this augmented MLMDP. Often hierarchical schemes require designing a pseudoreward function to encourage successful completion of a subtask. Here we also pick a set of reward functions over subtask states; however, the performance of our scheme is only weakly dependent on this choice: we require only that our chosen reward functions form a good basis for the set of subtasks that the higher layer will command. Any set of tasks which can linearly express the required space of reward structures specified by the higher level is suitable. In our experiments, we define $N^{l}_t$ tasks, one for each subtask state, and set each instantaneous reward to negative values on all but a single `goal' subtask state. Then the augmented MLMDP is $\tilde M^{l}=\left\langle \tilde S^l, \tilde P^{l}, q^{l}_i,  \left[Q^{l}_b ~ Q^{l}_t\right]\right\rangle$.

The higher level is itself an MLMDP $M^{l+1}=\left\langle S^l_t,P^{l+1}, q^{l+1}_i, Q^{l+1}_b\right\rangle$, defined not over the entire state space but just over the $N^{l}_t$ subtask states of the layer below. To construct this, we must compute an appropriate passive dynamics and reward structure. A natural definition for the passive dynamics is the probability of starting at one subtask state and terminating at another under the lower layer's passive dynamics,
\begin{eqnarray}
	P^{l+1}_i & = & \tilde P^{l}_t(I-\tilde P^{l}_i)^{-1}\tilde P^{{l}^T}_t, \label{eq_phi}\\
	P^{l+1}_b & = & \tilde P^{l}_b(I-\tilde P^{l}_i)^{-1}\tilde P^{{l}^T}_t. \label{eq_phb}
\end{eqnarray} 

In this way, the higher-level LMDP will incorporate the transition constraints from the layer below.

The interior-state reward structure can be similarly defined, as the reward accrued under the passive dynamics from the layer below. However for simplicity in our implementation, we simply set small negative rewards on all internal states.

Hence, from a base MLMDP $M^l$ and subtask transition matrix $P^l_t$, the above construction yields an augmented MLDMP $\tilde M^l$ at the same layer and unaugmented MLDMP $M^{l+1}$ at the next higher layer. This procedure may be iterated to form a deep stack of MLMDPs $\left\{ \tilde M^1,\tilde M^2,\cdots, \tilde M^{D-1}, M^D \right\},$ where all but the highest is augmented with subtask states. The key choice for the designer is $P^{l}_t$, the transition structure from internal states to subtask states for each layer. Through Eqns.~\req{eq_phi}-\req{eq_phb}, this matrix specifies the state abstraction used in the next higher layer. Fig.~\ref{hierarchy_overview} illustrates this scheme for an example of navigation through a 1D corridor.

\subsection{Instantaneous rewards and task blends: Communication between layers}

Bidirectional communication between layers happens via subtask states and their instantaneous rewards. The higher layer sets the instantaneous rewards over subtask states for the lower layer; and the lower layer signals that it has completed a subtask and needs new guidance from the higher layer by transitioning to a subtask state.

In particular, suppose we have solved a higher-level MLMDP using any method we like, yielding the optimal action $a^{l+1}$. This will make transitions to some states more likely than they would be under the passive dynamics, indicating that they are more attractive than usual for the current task. It will make other transitions less likely than the passive dynamics, indicating that transitions to these states should be avoided. We therefore define the instantaneous rewards for the subtask states at level $l$ to be proportional to the difference between controlled and passive dynamics at the higher level $l+1$, 
\begin{equation}
r_t^l \propto a^{l+1}_i(\cdot|s) - p^{l+1}_i(\cdot|s). \label{eqn_reward_inpainting}
\end{equation}
This effectively inpaints extra rewards for the lower layer, indicating which subtask states are desirable from the perspective of the higher layer.

The lower layer MLMDP then uses its basis of tasks to determine a task weighting $w^l$ which will optimally achieve the reward structure $r^l_t$ specified by the higher layer by solving \req{multitask_blend}. The reward structure specified by the higher layer may not correspond to any one task learned by the lower layer, but it will nonetheless be performed well by forming a concurrent blend of many different tasks and leveraging the compositionality afforded by the LMDP framework.

The true state of the agent is represented in the base level $\tilde M^1$, and next states are drawn from the controlled transition distribution. If the next state is an interior state, one unit of time passes and the state is updated as usual. If the next state is a subtask state, the next layer of the hierarchy is accessed at its corresponding state; no `real' time passes during this transition. The higher level then draws its next state, and in so doing can access the next level of hierarchy by transitioning to one of its subtask states, and so on. At some point, a level will elect not to access a subtask state; it then transmits its desired rewards from Eqn.~\req{eqn_reward_inpainting} to the layer below it. The lower layer then solves its multitask LMDP problem to compute its own optimal actions, and the process continues down to the lowest layer $\tilde M^1$ which, after updating its optimal actions, again draws a transition. Lastly, if the next state is a terminal boundary state, the layer terminates itself. This corresponds to a level of the hierarchy determining that it no longer has useful information to convey. Terminating a layer disallows future transitions from the lower layer to its subtask states, and corresponds to inpainting infinite negative rewards onto the lower level subtask states.

\section{Computational complexity advantages of hierarchical decomposition}
\label{sec_computational_complexity}

To concretely illustrate the value of hierarchy, consider navigation through a 1D ring of $N$ states, where the agent must perform $N$ different tasks corresponding to navigating to each particular state. We take the passive dynamics to be local (a nonzero probability of transitioning just to adjacent states in the ring, or remaining still). In one step of Z iteration (Eqn.~5), the optimal value function progresses at best $O(1)$ states per iteration because of the local passive dynamics (see \cite{Precup1998} for a similar argument). It therefore requires $O(N)$ iterations in a flat implementation for a useful value function signal to arrive at the furthest point in the ring for each task. As there are $N$ tasks, the flat implementation requires $O(N^2)$ iterations to learn all of them.

Instead suppose that we construct a hierarchy by placing a subtask every $M = \log N$ states, and do this recursively to form $D$ layers. The recursion terminates when $N/M^D \approx 1$, yielding $D\approx\log_M N$. With the correct higher level policy sequencing subtasks, each policy at a given layer only needs to learn to navigate between adjacent subtasks, which are no more than $M$ states apart. Hence Z iteration can be terminated after $O(M)$ iterations. At level $l=1,2,\cdots,D$ of the hierarchy, there are $N/M^l$ subtasks, and $N/M^{l-1}$ boundary reward tasks to learn. Overall this yields
\beq
\sum_{l=1}^{D}M\left(\frac{N}{M^{l}} + \frac{N}{M^{l-1}}\right)  =  N(1+M)\frac{1-(1/M)^{D }}{ 1-1/M} \leq  N(1+M) \approx  O(N\log N)
\eeq
total iterations. A similar analysis shows that this advantage holds for memory requirements as well. The flat scheme requires $O(N^2)$ nonzero elements of $Z$ to encode all tasks, while the hierarchical scheme requires only $O(N\log N)$. Hence hierarchical decomposition can yield qualitatively more efficient solutions and resource requirements, reminiscent of theoretical results obtained for perceptual deep learning \cite{Bengio2009,Bottou2007}. We note that this advantage only occurs in the multitask setting: the flat scheme can learn one specific task in time $O(N)$. Hence hierarchy is beneficial when performing an ensemble of tasks, due to the reuse of component policies across many tasks (see also \cite{Solway2014}).

\section{Experimental Demonstration}

\begin{figure}
\begin{center}
\includegraphics[width=\textwidth]{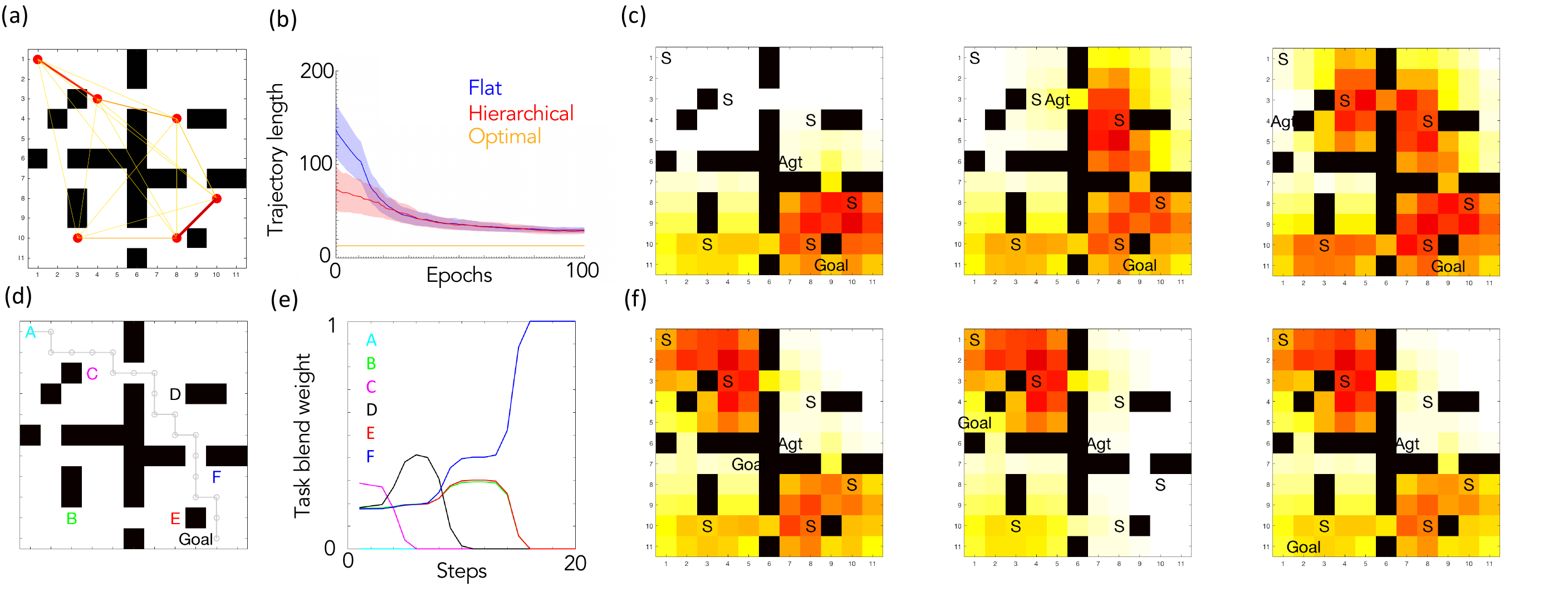}
\end{center}
\caption{ Experimental demonstration. (a) Four room domain with subtask locations marked as red dots, and derived higher-level passive dynamics shown as weighted links between subtasks. (b) Trajectory length over learning epochs with and without the help of the hierarchy. (c) Desirability function over states as agent moves through environment, showing the effect of reward inpainting from the hierarchy. (d) Sample trajectory (grey line) from upper left to goal location in bottom right. (e) Evolution of distributed task weights on each subtask location over the course of the trajectory in panel (d). (f) Desirability function over states as goal location moves. Higher layers inpaint rewards into subtasks that will move the agent nearer the goal.}
\label{fig_experimental_results}
\end{figure}

To illustrate the operation of our scheme, we apply it to a 2D grid-world `rooms' domain (Fig.~\ref{fig_experimental_results}A). The agent is required to navigate to a goal location in one of four rooms, and can move in the four cardinal directions or remain still. 

To build a hierarchy, we place several higher layer subtask goal locations throughout the domain (Fig.~\ref{fig_experimental_results}(a), red dots). The inferred passive dynamics for the higher layer MLMDP is shown in Fig.~\ref{fig_experimental_results}(a) as weighted lines between subtask states. The higher layer passive dynamics conform to the structure of the problem, with the probability of transition between higher layer states roughly proportional to the distance between those states at the lower layer.

As a basic demonstration that the hierarchy conveys useful information, Fig. \ref{fig_experimental_results}(b) shows Z-learning curves for the base layer policy with and without an omniscient hierarchy (i.e., one preinitialized using Z-iteration). From the start of learning, the hierarchy is able to drive the agent to the vicinity of the goal.

Fig.~\ref{fig_experimental_results}(d-e) visualizes the evolving distributed task representation commanded by the higher layer to the lower layer over the course of a trajectory. At each time point, several subtasks have nonzero weight, highlighting the concurrent execution in the system. Fig.~\ref{fig_experimental_results}(c) shows the composite desirability function resulting from the concurrent task blend for different agent locations. Finally, Fig.~\ref{fig_experimental_results}(f) highlights the multitasking ability of the system, showing the composite desirability function as the goal location is moved. The leftmost panel, for instance, shows that rewards are painted concurrently into the upper left and bottom right rooms, as these are both equidistant to the goal.

\section{Conclusion}
The Multitask LMDP module provides a novel approach to control hierarchies, based on a distributed representation of tasks and parallel execution. Rather than learn to perform one task or even a fixed collection of tasks, it exploits the compositionality provided by linearly solvable Markov decision processes to perform an infinite space of task blends optimally. Stacking the module yields a deep hierarchy abstracted in state space and time, with the potential for qualitative efficiency improvements. 

While a variety of sophisticated reinforcement learning methods have made use of deep networks as capable function approximators \cite{Mnih2015,Lillicrap2015,Levine2016End-to-EndPolicies}, in this work we have sought to transfer some of the underlying intuitions, such as parallel distributed representations and stackable modules, to the control setting. In the future this may allow other elements of the deep learning toolkit to be brought to bear in this setting, most notably gradient-based learning of the subtask structure itself.

{\bf Acknowledgements} We acknowledge support from the National Research Foundation of South Africa, and the Swartz Foundation.


\end{document}